\documentclass[letterpaper, 10 pt, journal, twoside]{IEEEtran}

\usepackage{preamble}

\title{Estimating Visibility from Alternate Perspectives for \\ Motion Planning with Occlusions}

\urldef{\smith}\url{stephen.smith@uwaterloo.ca}
\urldef{\gilhuly}\url{barry.gilhuly@uwaterloo.ca}
\urldef{\sadeghi}\url{a6sadegh@uwaterloo.ca}

\author{Barry Gilhuly, Armin Sadeghi, and Stephen L.\ Smith
    \thanks{This research is partially supported by the Natural Sciences and Engineering Research Council of Canada (NSERC). }
    \thanks{The authors are with the Department of Electrical and Computer Engineering, University of Waterloo, Waterloo ON, N2L 3G1 Canada (\gilhuly, \sadeghi, \smith)}
    \thanks{Digital Object Identifier (DOI): see top of this page.} 
}

\makeatletter
\def\ps@IEEEtitlepagestyle{%
  \def\@oddfoot{\mycopyrightnotice}%
  \def\@oddhead{\hbox{}\@IEEEheaderstyle\leftmark\hfil\thepage}\relax
  \def\@evenhead{\@IEEEheaderstyle\thepage\hfil\leftmark\hbox{}}\relax
  \def\@evenfoot{}%
}
\def\mycopyrightnotice{%
  \begin{minipage}{\textwidth}
  \centering \scriptsize
    This work has been submitted to the IEEE for possible publication. Copyright may be transferred without notice, after which this version may no longer be accessible.  
  \end{minipage}
}
\makeatother

\begin{document}

\newpage
\maketitle

\begin{abstract}
    Visibility is a crucial aspect of planning and control of autonomous vehicles (AV), particularly when navigating environments with occlusions. However, when an AV follows a trajectory with multiple occlusions, existing methods evaluate each occlusion individually, calculate a visibility cost for each, and rely on the planner to minimize the overall cost. This can result in conflicting priorities for the planner, as individual occlusion costs may appear to be in opposition. We solve this problem by creating an alternate perspective cost map that allows for an aggregate view of the occlusions in the environment. The value of each cell on the cost map is a measure of the amount of visual information that the vehicle can gain about the environment by visiting that location. Our proposed method identifies observation locations and occlusion targets drawn from both map data and sensor data.  We show how to estimate an alternate perspective for each observation location and then combine all estimates into a single alternate perspective cost map for motion planning.    
\end{abstract}


\section{Introduction}

\IEEEPARstart{V}{isibility}, a representation of the proportion of the environment that is measurable by on-board sensors, is an important consideration in planning the path of an autonomous vehicle (AV).%
Collision metrics, such as Time-To-Collision (TTC)~\cite{schwarz2014computing}, Time-To-React (TTR)~\cite{nager2019lies}, and others~\cite{lin2023commonroad} depend on knowing the distance from other agents in the environment. Potential agents that are not captured by sensors, such as those hidden by occlusions, are not visible to the vehicle and instead must be estimated.

\begin{figure}
    \centering
    \begin{subfigure}{0.4\linewidth}
        \centering
        \includegraphics[width=0.8\linewidth]{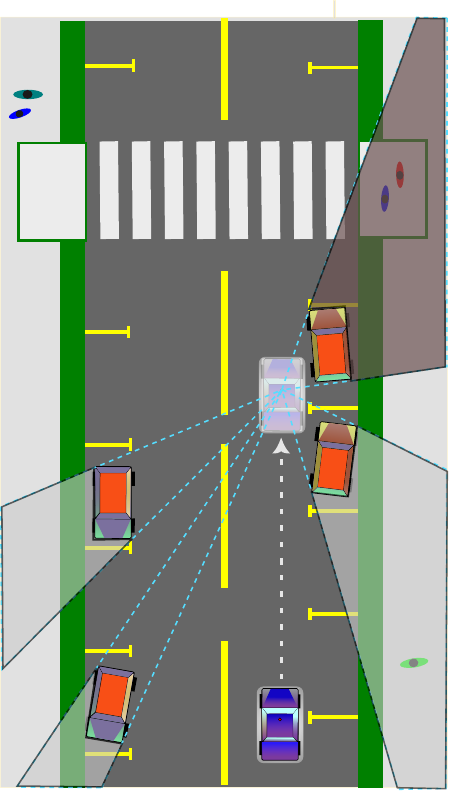}
        \caption{An occluded crosswalk}
        \label{fig:visibility-demo-occluded}
    \end{subfigure}
    \begin{subfigure}{0.4\linewidth}
        \centering
        \includegraphics[width=0.8\linewidth]{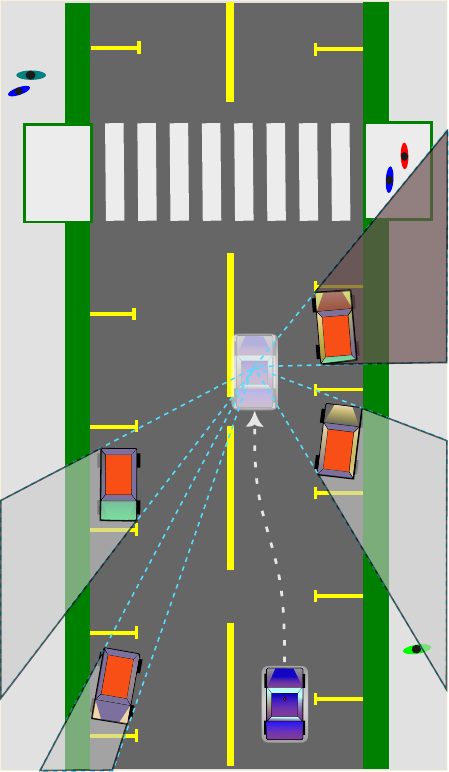}
        \caption{Pedestrians revealed}
        \label{fig:visibility-demo-revealed}
    \end{subfigure}
    \caption{An example of two possible trajectories for an autonomous vehicle with occlusions highlighted. In (\protect\subref{fig:visibility-demo-occluded}), the AV maintains the nominal trajectory resulting in an occlusion that hides two pedestrians (darker shadow in upper right).  In (\protect\subref{fig:visibility-demo-revealed}), the AV takes a position on the left side, providing earlier perception of the crosswalk entry.}
    \label{fig:visibility-demo}
\end{figure}

Normally, as the AV moves through its environment following its nominal trajectory in the center of the lane, obscured areas eventually become revealed and hidden agents are made visible. 
However, in many cases, it is possible that some critical subjects are not revealed until the AV gets too close (\eg Figure~\ref{fig:visibility-demo}, left). In such cases, it may be helpful to preemptively shift the AV trajectory to reveal the occluded area earlier (\eg Figure~\ref{fig:visibility-demo}, right), thus giving the AV more time to determine if it needs to slow or stop.

We propose a method to estimate the informative value of positions in an environment and to store that information in a grid-based cost map, which we refer to as an \emph{Alternate Perspective Cost Map} (APCM). Those positions along a trajectory that provide more or earlier views into an occluded area are more valuable to visit and are assigned a positive reward.  Alternatively, positions that delay the reveal of hidden areas offer no advantage to visit, and thus have no reward. 

The APCM can then be used to refine the nominal trajectory so that the AV visits more valuable positions within its lane, thus maximizing its view of the upcoming environment and improving visibility estimates for areas with occlusions. 

As an example, consider a situation where the AV has the choice of lane position, as shown in Figure~\ref{fig:visibility-demo}. If the AV follows the nominal trajectory it will remain in the center of its lane, but in this central position the entry to the crosswalk remains occluded due to a parked car.  The crosswalk entry is only revealed when the AV is too close to do anything but slow in anticipation. However, if the AV moves toward the left side of the lane, any pedestrians waiting to cross are revealed earlier, allowing the AV to maintain speed if no pedestrians are revealed, or slow down if necessary.

Our focus is on environments where high-definition maps are available.  High-definition maps allow planning based on occlusions that may be outside the current sensor range, enabling the prediction to be effective over a longer horizon.

\subsection{Previous Work}

Planning to manage the regions of the environment visible to an autonomous vehicle is an important aspect of planning and control~\cite{adelberger2022topology,tariq2022vehicle}.  The authors of~\cite{hubmann2019pomdp} use a POMDP model to simulate how the field of view changes as the autonomous vehicle moves. 

Research on maximizing visibility has previously appeared in~\cite{higgins2022model,andersen2019trajectory,li2021occlusion,christianos2023planning,gilhuly2022looking}. In~\cite{higgins2022model}, the authors reduce the expense of precise calculations using a function that closely approximates the space occluded behind a cylindrical obstacle. Although this method of fast approximations provides continuous cost estimates, it relies on representing all areas as circles, thus limiting the ability to represent more realistic structures found in an urban environment. Furthermore,~\cite{higgins2022model} does not consider the uncertainty that can occur when detecting obstacles. 
Another approach uses a scaled multiple of the incidence angle as a visibility reward in the optimization cost function~\cite{andersen2019trajectory}. However, this method does not consider environments where multiple occlusions must be processed simultaneously. 
In~\cite{li2021occlusion}, by defining the region where a pedestrian may suddenly appear, the AV path can be modified to maintain a safe distance. Alternatively, the AV may plan in anticipation of the trajectories that agents  emerging from an occlusion could follow~\cite{christianos2023planning}.  In contrast to our approach, both~\cite{li2021occlusion} and~\cite{christianos2023planning} focus on avoiding occlusions rather than seeking an earlier perception of occluded regions.  

Cost maps are used in robotics for a wide array of purposes. Occupancy grid maps (OGM)~\cite{thrun2003learning,plebe2021occupancy,edmonds2021optimal,kanai2022cooperative} maintain belief in the state of the robot environment, giving each grid cell a probabilistic occupancy value.  This allows the AV to track and avoid locations that are likely to cause a collision. Traversability cost maps\cite{fan2021learning} estimate the risk or resource cost to the robot when moving through certain spaces in the environment, but are static in nature. In~\cite{rajendran2021human, mainprice2011planning}, visibility cost maps allow the robot to consider the visual perspective of the human in the motion planning sequence. 

In our own previous work, we showed that estimating future visibility can lead to more informative trajectories~\cite{gilhuly2022looking}. Building on our previous work, here we construct an online cost map from visibility estimates that offers new benefits.  Creating the APCM allows for efficient evaluation of multiple trajectories, reducing perception estimation at each position to a single lookup operation.
Contrary to the previous work where the trajectories determined a limited set of observation locations, in this work we evaluate the entire discretized AV planning space.  With this change, the observations no longer depend on the quality of the chosen trajectories. 
The resulting cost map is also in an easily consumable format that is readily adaptable to standard optimization planners, as we demonstrate in the simulation.   

An element of visibility planning that has not been explored well is operation in cluttered environments. Cluttered environments occur when there are many obstacles near the AV trajectory, each generating an occlusion, \eg driving on a two-lane road with cars parked on both sides.  Existing planners experience conflicting demands that cannot be resolved by altering the trajectory, as moving away from occlusions on one side to decrease costs increases costs on the other side.  The resulting behavior can be undesirable, as the planner either remains on the nominal trajectory or moves in unexpected ways.  
In this paper, we address this issue through the APCM. Construction of the cost map yields a perspective reward for visiting a location relative to the other locations along the AV trajectory.  This, in turn, allows the planner to make more informed decisions about its trajectory.

\subsection{Contributions}

This paper contributes the following.  
First, we introduce the alternate perspective cost map (APCM): 
given a current AV location and the regions occluded from the AV's view, an  APCM is a discretization of the environment, where each cell contains an estimate how much of the occluded area would be visible if the AV were able to take an observation from that location (\ie that alternate perspective). 
Second, we show that the APCM allows the AV to plan trajectories that reveal occluded regions earlier than would otherwise occur when following the nominal trajectory.  In addition, APCM enables the AV to predict future visibility along planned trajectories.
Third, by integrating our cost map into a Model Predictive Control (MPC~\cite{rawlings2000tutorial}) based planner, we demonstrate an improved response when occlusions are present on both sides of the AV when compared to state-of-the-art visibility-based planners.
Finally, we propose a GPU-based implementation that allows for operation in real time.

\section{Problem Statement}
In this section, we provide background information and context before introducing the main problem.

\emph{Notation:} We use the calligraphic font \eg $\mathcal{U}$, to denote sets, and the roman font to denote variables. Lowercase bold letters, \eg $\mathbf{x}$, represent vectors, and the standard font, \eg $y$, represents scalars such as vector components. Matrices are generally represented by uppercase bold letters, \eg $\mathbf{Q}$; however, we represent an $N\times N$ cost map $\mathbf{P}$ as a function $\mathbf{P}: N\times N \to \real$, where the value of cell $c\in N\times N$ is $\mathbf{P}(c)$.  

An autonomous vehicle in an environment $\mathcal{E} \subseteq \mathbb{R}^2$ has the current position $\mathbf{x}_t$ and a planned trajectory $x_{\textsc{traj}} = (\mathbf{x}_{t+1}, \ldots, \mathbf{x}_{t+T})$.  With the exception of the ego vehicle and hidden pedestrians, other elements of the environment are assumed to be static.
The AV has access to a map $\mathbf{M}^{\textsc{HD}}$ of the environment containing occupancy information, possibly reconstructed from previous experience or retrieved from a high definition reference.  The map is accurate with respect to fixed obstacles and surfaces (\eg buildings, lampposts, and roads), but does not contain information on mobile elements (\eg  parked cars, pedestrians, and on-road vehicles).

At each time step from $0, \ldots, T$, the AV takes a sensor measurement and produces an OGM $\mathbf{M}_t^{\textsc{ogm}}$.   We assume that each cell of $\mathbf{M}_t^{\textsc{ogm}}$ contains a probabilistic estimate of occupancy and is independent of its neighbors~\cite{thrun2005probabilistic}.   We assume that the sensor readings are without error.

We define the area of $\mathcal{E}$ that is \emph{uncertain} according to $\mathbf{M}_t^{\textsc{ogm}}$, obtained by thresholding $\mathbf{M}_t^{\textsc{ogm}}$, as $\mathcal{U}_t$.  These are the cells of $\mathbf{M}_t^{\textsc{ogm}}$ where the AV has limited or partial information only. Hence, the uncertainty usually arises from occlusions and represents the elements of $\mathcal{E}$ that the AV sensors could not measure.   We determine the type of agent, \eg $\mathcal{A} = \{ \texttt{car}, \texttt{pedestrian}, \texttt{bike}, \etc \}$, which could be located in each element of $\mathcal{U}_t$ according to $\mathbf{M}^{\textsc{HD}}$.  If $\mathbf{M}^{\textsc{HD}}$ is not available, then the worst-case agent behavior is considered. 

\begin{figure}
    \centering
    \begin{subfigure}[c]{0.85\linewidth}
        \centering
        \includegraphics[width=0.95\linewidth]{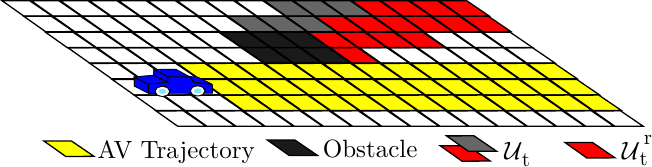}
        \caption{An occluded region with reachability marked.}
        \label{fig:simplified-perspective-a}
    \end{subfigure} \\
    \vspace{2mm}
    \begin{subfigure}[c]{0.85\linewidth}
        \centering
        \includegraphics[width=0.95\linewidth]{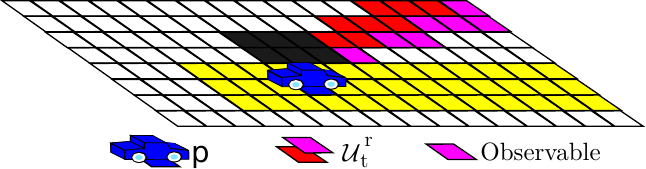}
        \caption{Approximately half of $\mathcal{U}_t^r$ is visible from $p$.}
        \label{fig:simplified-perspective-b}
    \end{subfigure} \\
    \vspace{2mm}
    \begin{subfigure}[c]{0.85\linewidth}
        \centering
        \includegraphics[width=0.95\linewidth]{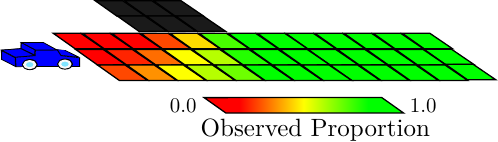}
        \caption{The completed cost map.  Cells ahead of the obstacle are expected to have a clear view of $\mathcal{U}_t^r$, while those approaching the AV are progressively occluded and thus have lower values.}
        \label{fig:simplified-perspective-c}
    \end{subfigure}
    \caption{Construction of the Perspective Cost Map.}
    \label{fig:simplified-visibility}
\end{figure}

We further define $\mathcal{U}_t^r \subseteq \mathcal{U}_t$ as a subset of the uncertain space where, if an agent is located at $\mathbf{z}\in \mathcal{U}_t^r$ at time $t$, it could reach the vehicle traveling along $x_{\textsc{traj}}$, while obeying its agent-specific maximum speed. See Figure~\ref{fig:simplified-visibility}.

More formally, we define the notion of \emph{reachability} in $\mathcal{E}$. 
Given the AV at position $\mathbf{x}_t$, and an agent $a$ at $\mathbf{z}$ with class-based maximum speed $V_{\text{max}}^a$, then the $n^{\text{th}}$ step of $x_{\textsc{traj}}$ is \emph{reachable} from $\mathbf{z}$ if
\begin{align} 
     \left\{\mathbf{z} \in \mathcal{U}_t \bigg | \frac{\|\mathbf{x}_{t+n} - \mathbf{z}\|}{n V_{\text{max}}^a} \leq 1 \right\}  \label{eqn:reachable-set}
\end{align}
Then, representing~\eqref{eqn:reachable-set} as $\texttt{reach}( \mathcal{U}_t, \mathbf{x} )$, the set of all points in $\mathcal{U}_t$ at time $t$ that can reach $x_{\textsc{traj}}$ is 
\begin{align} \label{eqn:reachable}
    \mathcal{U}_t^r = \bigcup_{n=1}^T \texttt{reach}( \mathcal{U}_t, \mathbf{x}_{t+n} ).
\end{align}

With this information, we construct the APCM $\mathbf{P}_t$.  We discretize the environment into an $N \times N$ grid with the same size and resolution as $\mathbf{M}_t^{\textsc{ogm}}$. Each cell $p \in \mathbf{P}_t$ stores a reward that reflects the reduction in uncertainty that would occur in $\mathcal{U}_t^r$  if the AV were to jump immediately to that cell at time $t$ and make an observation.  For example, in Figure~\ref{fig:simplified-perspective-b}, if the AV moved forward as shown, it would observe the purple cells, corresponding to $\approx \frac{1}{2}\mathcal{U}_t^r$. Thus, $\mathbf{P}_t$ stores a compact and efficient approximation of what the AV would see if it visited these cells at a future time step.
We enable robot planners to consider interactions with occlusions by representing perspectives in this way.
Although we do not exactly capture the true value, we will demonstrate the utility of our approach in the following sections.

\begin{remark} [Extension to Dynamic Occupancy]
    In this work, we assume that all perspective planning is performed as if the environment is frozen at time $t$, thus omitting dynamic elements.  The methods presented here can be extended to include a dynamic occupancy grid map in place of $\mathbf{M}_t^{\textsc{ogm}}$, leading to the creation of a spatiotemporal~\cite{mann_predicting_2022, yoon_estimation_2021, wang_research_2021, xin_enable_2021} perspective cost map.  We leave this development to future work.  
\end{remark}

We can now state our primary problem.
\begin{problem} \label{prb:cost-map}
    Given the occupancy grid $\mathbf{M}_t^{\textsc{ogm}}$ of size $N\times N$ and the map $\mathbf{M}^{\textsc{HD}}$, construct alternate perspective cost map $\mathbf{P}_t$, an $N\times N$ grid with each cell containing the expected observable fraction of $\mathcal{U}_t^r$ from that location.     
\end{problem}

In the remainder of this paper, we develop the construction of the APCM.  Then, we discuss a GPU implementation, exploiting the parallel nature of the problem.  Next, we present an application with a Model Predictive Control (MPC) based planner.  Finally, we show simulation results and comparisons with prior visibility maximization methods before concluding with some thoughts on future work.

\begin{remark}[Restricted to 2D]
    Note that in this work, the problem has been restricted to $\mathbb{R}^2$; however, the same methods can be extended to environments in $\mathbb{R}^3$.  
\end{remark}

\section{Construction of the APCM}
In this section, we develop a process for computing and updating the APCM $\mathbf{P}_t$. 

\subsection{Calculating the Reachable Occluded Space \texorpdfstring{$\mathcal{U}_t^r$}{Ur}}
Recall that $\mathcal{U}_t$ consists of those locations in the environment that are uncertain in $\mathbf{M}_t^{\textsc{ogm}}$.   We are interested in the cells in $\mathcal{U}_t$ for which the AV trajectory is reachable.
Given an intended AV trajectory consisting of  $x_{\textsc{traj}} = \{ \mathbf{x}_1, \mathbf{x}_2, \ldots, \mathbf{x}_T \}$, each with an expected arrival time $t = (1,2,\ldots,T)$,  let $\mathcal{A}_{n}^a$ represent the set of cells
%
that are \emph{reachable} as defined by~\eqref{eqn:reachable} 
at $\textbf{x}_n$ for each type of agent considered, $a \in \{ \textit{pedestrian}, \textit{bike}, \textit{car}, \etc \}$, parameterized by $t$ and the class-specific agent speed.  Then $\mathcal{U}_t^r$ is 
\begin{align}
    \mathcal{U}_t^r = \cup_{n=1}^T \left[ \cup_a \left( \mathcal{A}_{n}^a \cap \mathcal{U}_t \right) \right].
\end{align}
The set $\mathcal{U}_t^r$ can extend beyond $\mathbf{M}_t^{\textsc{ogm}}$ into the larger map $M_\textsc{HD}$. 
\begin{remark}[Importance of Thresholding]
 Reducing the size of $\mathcal{U}_t$ by thresholding $\mathbf{M}_t^{\textsc{ogm}}$ to include only cells with sufficient uncertainty reduces computational requirements and allows the planner to focus only on the areas where another agent has been seen or where an occlusion can pose a safety risk.   
\end{remark}

\subsection{Alternate Perspective Locations}
\begin{figure}
    \centering 
    \includegraphics[width=0.9\linewidth]{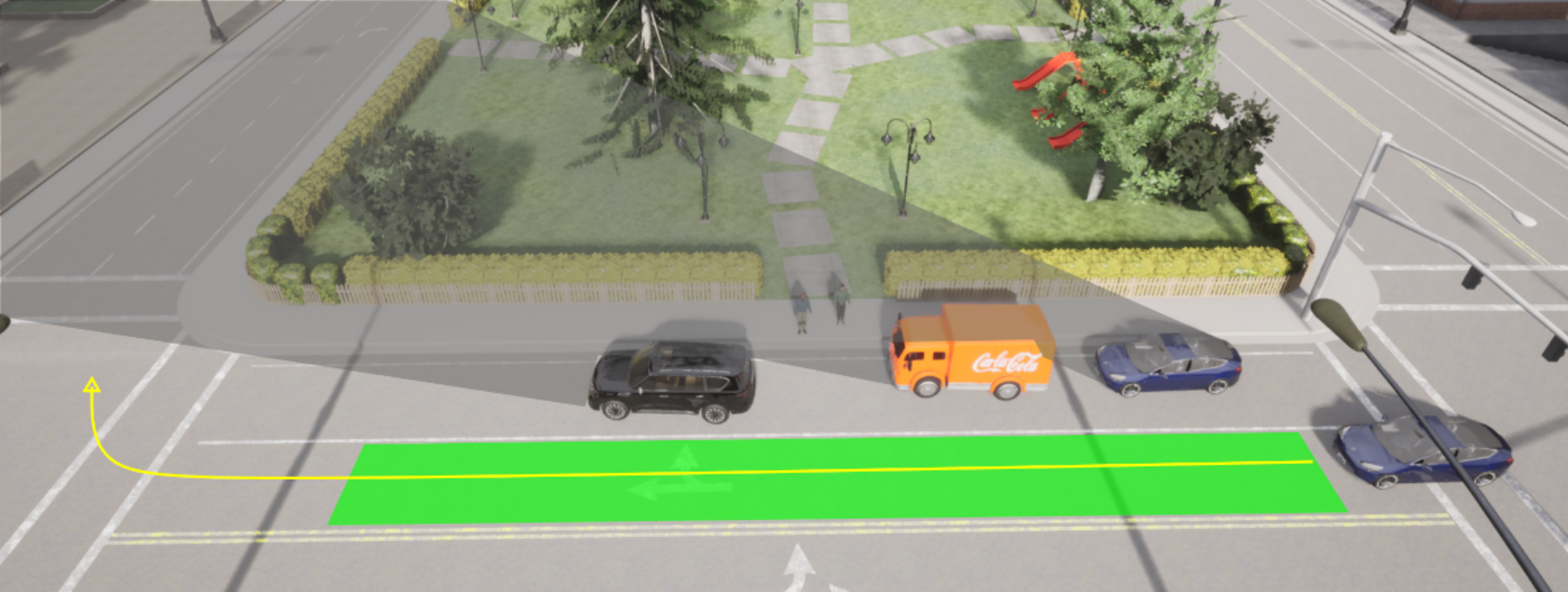}
    \caption{The observation locations are highlighted along the intended trajectory. Phantom pedestrians are inserted into the occluded space in front of the truck from where they could enter the roadway.}
    \label{fig:visibility-sources}
\end{figure}

We start the construction of the APCM by identifying the locations in the environment where alternative perspectives will be evaluated.  These locations are found along the AV trajectory that may be visited within the planning horizon.  We denote these locations as a set of cells $\mathcal{S}_t \subseteq \mathbf{P}_t$. Assuming that a traffic lane has a width of $d$ m, then cells $s \in\mathcal{S}_t$ are any cells that are along or within a distance $\nicefrac{d}{2}$ of the nominal trajectory and can be reached within the planning horizon (Figure~\ref{fig:visibility-sources}).   All other cells of $\mathbf{P}_t$ have a perspective value of zero.  
Although not developed here, $\mathcal{S}_t$ can be based on multiple nominal trajectories, allowing visibility prediction to contribute to more complex maneuvers, such as lane changes.

\subsection{Calculating Alternate Perspectives} \label{sec:calc-ap}
We assess the probability of observing a cell $u$ in the uncertain region $\mathcal{U}_t^r$ from the cell $s$ by estimating the probability of a ray traveling from $s$ to $u$, given $\mathbf{M}_t^{\textsc{ogm}}$ and $\mathbf{M}^{\textsc{hd}}$.  

First, we create a general occupancy map $\mathbf{M}_t^{\textsc{occ}}$ by overlaying $\mathbf{M}_t^{\textsc{ogm}}$ on $\mathbf{M}^{\textsc{hd}}$.  This creates a single map with current occupancy information local to the AV and one/zero occupancy in regions outside the sensor range.
Let $\mathcal{R}$ represent the set of cells in $\mathbf{M}_t^{\textsc{occ}}$ that are intersected by the ray from the source cell $s$ to the target cell $u$, without including the end points.  A ray passing through any cell $c$ in $\mathcal{R}$ is predicted to be blocked by the probability of cell occupancy. The probability of observing the target cell $u$ from $s$ is
\begin{align} \label{eqn:visibility-prob}
    \pb_{\textsc{view}}(u) = \prod_{c \in \mathcal{R}} \big(1 - \mathbf{M}_t^{\textsc{occ}}(c)\big). 
\end{align}
The alternate perspective value for cell $s$, denoted $s^p$, is the sum of~\eqref{eqn:visibility-prob} for all $u$ in $\mathcal{U}_t^r$ divided by the size of $\mathcal{U}_t^r$, written
\begin{align}
    s^p = \frac{\sum_{u \in \mathcal{U}_t^r} \pb_{\textsc{view}}(u)}{|\mathcal{U}_t^r|}.     \label{eqn:value-of-visibility}
\end{align}
Thus, if all $\mathcal{U}_t^r$ is observable, then $s^p = 1$. 
Next, the perspective values are scaled and normalized to a range of $[0,1]$.  This removes any bias that occurs versus cells outside $\mathcal{S}_t$ for which there is no estimate.  Note that if the perspective values are uniform, \ie all cells have the same value, then after normalization all $s^p = 0$.  We update the APCM, $\mathbf{P}_t(s) = s^p$. We have now calculated the estimate of the relative perspective value of visiting each $s$ at time $t$.  

\begin{remark}[Proportion of $\mathcal{U}_t^r$ observed]
    Since each source cell only stores an estimate of the proportion of\hspace{0.5em}$\mathcal{U}_t^r$ that will be observable, there may be some overlap in the perspective value between cells, depending on their separation.  Therefore, given two observation locations $s_i$ and $s_j$, the proportion of\hspace{0.5em}$\mathcal{U}_t^r$ observed if both $s_{i}$ and $s_{j}$ are visited has a lower bound of $\max( |s_{i}^p|, |s_{j}^p| )$ when $s_{i}$ and $s_{j}$ are close together and share similar views, and an upper bound of $|s_{i}^p| + |s_{j}^p|$ when the visible components of\hspace{0.5em}$\mathcal{U}_t^r$ from each position are disjoint.
\end{remark}

The creation and update of the APCM $\mathbf{P}_t$ solves Problem~\ref{prb:cost-map}.  In the next section, we provide an application of an optimization controller that uses $\mathbf{P}_t$ to plan trajectories that maximize the future perspective view of the AV.

\section{GPU Implementation}

The APCM update requires making perception estimates of $M$ target points from $K$ observation locations.  Importantly, each perception estimate is independent, allowing us to take advantage of parallel processing on the GPU.   

\subsection{Computational Complexity}
We use Bresenham's algorithm~\cite{bresenham1977linear} to cast a ray between cells $s$ and $u$ as described in~\ref{sec:calc-ap}, which consists of a single loop of at most $\nicefrac{N}{2}$ steps on an $N \times N$ grid.  Since there are a maximum of $N \times N$ cells in $\mathcal{U}_t^r$, the complexity of building a single source observation point is $O(N^3)$.  Assuming $K$ observation cells, the total complexity is $O(KN^3)$.  Although it is possible for the number of observation points to be equal to $N \times N$, we have restricted the observation locations making $K \ll N \times N$.

\subsection{Algorithm Details}
The visibility probability of each ray is calculated independently, making this problem an excellent candidate for a GPU implementation\footnote{GPU implementation can be found at~\url{https://github.com/idlebear/APCM}}. For reference, there are many GPU-based implementations to build cost maps~\cite{stepanasOHMgpu, min2023octomap, castro2023does, williams2017information}.  Similarly to previous work, we exploit the parallel architecture of the visibility problem to enable fast computation on a GPU.  

The procedure for computing the APCM is given in Algorithm~\ref{alg:visibility}.  The inputs are a list of $K$ observation sources (\texttt{srcs}), $M$ targets (\texttt{trgs}), and a probabilistic occupancy map (\texttt{occ}).  First, the memory is allocated to contain all $K \times N$ alternate perspective predictions and $K$ aggregate perspective values (lines~\ref{alg:allocate}).  Next, for each of the projected rays, a thread is started (lines~\ref{alg:loop-start}-\ref{alg:loop-end}), and the ray casting algorithm is called (line~\ref{alg:bresenham}).  Once all rays are calculated, the visibility values are summed ($K$ threads, lines~\ref{alg:sum-loop-start}-\ref{alg:sum-loop-end}), normalized, and the visibility result is returned (line~\ref{alg:result}).

\begin{algorithm}[!t]
\small
\caption{The APCM update from a set of sources to a set of targets}
\label{alg:visibility}
\begin{algorithmic}[1]
\Procedure{UpdateAPCM}{ \texttt{srcs}, \texttt{trgs}, \texttt{occ} }
\State $K \gets \textbf{sizeof}(\texttt{srcs}), ~M \gets \textbf{sizeof}(\texttt{trgs})$
\State $\texttt{ray} \gets \texttt{Zeros}(K,M),~\texttt{vis} \gets \texttt{Zeros}(K)$ \label{alg:allocate}
\For {$k$ \textbf{in} $0\ldots K-1$} \label{alg:loop-start} \Comment{Start thread for each ray}
    \For {$m$ \textbf{in} $0\ldots M-1$}
        \State $s \gets \texttt{srcs}[k],~t \gets \texttt{trgs}[m] $
        \State $v \gets \textbf{cast\_ray}( s, t, occ )$   \label{alg:bresenham}
        \State $\texttt{ray}[k][m] \gets v$
    \EndFor
\EndFor \label{alg:loop-end}
\For {$k$ \textbf{in} $0\ldots K-1$} \label{alg:sum-loop-start}  \Comment{Sum visibility}
    \State  $\texttt{vis}[k] \gets 0$
    \For {$m$ \textbf{in} $0\ldots M-1$}
        \State  $\texttt{vis}[k] \gets \texttt{vis}[k] + \texttt{ray}[k][m]$ 
    \EndFor
\EndFor 
\For {$k$ \textbf{in} $0\ldots K-1$}   \Comment{Equation~\eqref{eqn:value-of-visibility}}
    \State  $\texttt{vis}[k] \gets \texttt{vis}[k] / M$
\EndFor \label{alg:sum-loop-end}
\State \textbf{return} $\texttt{vis}$  \label{alg:result}
\EndProcedure
\end{algorithmic}
\normalsize
\end{algorithm}

\section{Application to MPC}
In this section, we consider the problem of an AV traveling along a trajectory lined with parked cars which may conceal pedestrians, as shown in Figure~\ref{fig:visibility-sources}.    We use these parked cars and hidden pedestrian scenarios because they are a common occurrence in urban driving and are frequently discussed in the existing literature~\cite{li2021occlusion,wang2023occlusion,andersen2019trajectory,zhang2021improved,van2023overcoming}.   We formulate a solution to this problem using a Model Predictive Path Integral (MPPI~\cite{williams2017model}) controller, an MPC variant, modified to include visibility cost.  

\subsection{Vehicle Model} \label{subsec:vehicle-model}
We use a second-order bicycle kinematic model for vehicle simulation and control.  The vehicle state, $\mathbf{x}$, consists of $x,y$ position, velocity, and orientation,
\begin{align}
    \mathbf{x} = \begin{bmatrix}
        x, y, v, \theta
    \end{bmatrix}'.
\end{align}
The control inputs are acceleration and steering angle, $\mathbf{u}_t = \begin{bmatrix}
    a, \delta
\end{bmatrix}'$, and the vehicle model assumes tracking based on the rear axle of the vehicle.  For a vehicle with a wheelbase $L$, the state update is
\begin{align} \label{eqn:continuous-model}
    \dot{\mathbf{x}} = 
    \begin{bmatrix}
        \dot{x} \\
        \dot{y} \\
        \dot{v}  \\
        \dot{\theta} \\
    \end{bmatrix} = 
    \begin{bmatrix}
        v \cos{(\theta)} \\
        v \sin{(\theta)} \\
        a \\
        v \tan{(\delta)} / L\\
    \end{bmatrix}.
\end{align}
The continuous model~\eqref{eqn:continuous-model} is discretized using a fourth-order Runge-Kutta method (RK4)~\cite{butcher1996history}.

\subsection{The Optimal Control Problem} \label{subsec:ocp}
MPPI is a sampling-based variant of MPC and is used in this work to solve the optimal control problem (OCP) and, as such, is a straightforward integration with APCM.   We define the diagonal matrices $\mathbf{Q}$ and $\mathbf{Q}_\mathbf{f}$ for the state error costs over the planning horizon and the final term, respectively. A diagonal matrix, $\mathbf{R}$ represents the control error cost.  
Obstacles in the environment are represented by $\mathcal{O}$, each having a position and a bounding polygon.
Let $J_{\textsc{vis}}(\mathbf{x}_t)$ be a function to calculate the visibility cost to be included in the OCP, further defined in Section~\ref{sec:baselines}.  
Then, given a reference trajectory $\mathbf{x}_{1:T}^r$ and a reference control $\mathbf{u}_{1:T}^r$, we define the OCP as follows:
\begin{align}
    \textbf{min } &\|\mathbf{x}_T-\mathbf{x}_T^r\|_{\mathbf{Q}_\mathbf{f}}^2 + \nonumber \\ 
    & ~\sum_{t=1}^T \bigg [ \|\mathbf{x}_t-\mathbf{x}_t^r\|_\mathbf{Q}^2 + \|\mathbf{u}_t-\mathbf{u}_t^r\|_\mathbf{R}^2 + J_{\textsc{vis}}(\mathbf{x}_t, \mathbf{u}_t) \bigg ] \label{eqn:ocp-2} \\ 
    \mathbf{s.t.\;} & \mathbf{x}_{t+1} = f(\mathbf{x}_t,\mathbf{u}_t), t \in \{0\ldots T-1\} \label{eqn:ocp-3}\\
    & \texttt{dist}(\mathbf{o}_i, \mathbf{x}_t) > r_{\textsc{safe}}~\forall o_i \in \mathcal{O}, t \in \{1\ldots T\}. \label{eqn:ocp-4}
\end{align}
The term $\|\cdot\|_A^2$ indicates a quadratic of the form $\mathbf{x}'A \mathbf{x}$.  The vehicle model from Section~\ref{subsec:vehicle-model} forms the constraint~\eqref{eqn:ocp-3} and avoidance of all obstacles in $\mathcal{O}$ forms the constraint~\eqref{eqn:ocp-4}.
Our implementation of the MPPI planner limits costs to tracking, avoiding static obstacles in the environment, and perception.  The resulting trajectory is then evaluated by a safety controller that checks for potential collisions due to phantom agents, such as pedestrians.  If a potential collision is detected, the ego car is slowed.  We anticipate that the APCM could be integrated into a method such as~\cite{mohamed2022autonomous} resulting in a controller that handles dynamic obstacles while incorporating perception into planning, but leave this for future work.

\subsection{Baseline Methods} \label{sec:baselines}
For comparison, we implemented two of the state-of-the-art visibility maximization methods: Higgins and Bezzo~\cite{higgins2022model} and Andersen et al.~\cite{andersen2019trajectory}.  Each method implements $J_{\textsc{vis}}(\mathbf{x}_t)$ in~\eqref{eqn:ocp-2} in order to include a visibility cost in the OCP.

For all simulations, the OCP is solved using MPPI as described in Section~\ref{subsec:ocp}.  Control outputs from the OCP are evaluated for safety using a TTC metric.  If the safety check fails, the requested input control acceleration is reduced to a safe level or braking is applied.  

\subsubsection{Higgins and Bezzo~\texorpdfstring{\cite{higgins2022model}}{Higgins}}

The authors of~\cite{higgins2022model} apply visibility maximization to the problem of path planning for small robots.  We have extended their approach to autonomous vehicles.  As in~\cite{higgins2022model}, we approximate occlusions by fitting a circle over each sensed obstacle to calculate the area occluded.  Given a set of obstacles $\mathcal{O}$, for each time step, the visibility cost is
\begin{align} \label{eqn:higgins-obstacle-cost}
J_{\textsc{vis}}(\mathbf{x}_t) = M \sum_{o}^{\mathcal{O}} \left[\ln\big[ 1 + \exp\big( \frac{r_o}{d_o}(r_{\textsc{fov}}^2 - d_o^2) \big) \big] \right]^2,
\end{align}
where $d_o = \texttt{dist}(\mathbf{o}, \mathbf{x}_t)$ is the distance from the AV to obstacle $o$, $r_o$ is the fitting circle radius, $r_{\textsc{fov}}$ is the radius of the AV sensors, and $M$ is a scaling factor.  In our implementation, if $r_{fov}^2 \gg d_o^2$, then each term of~\eqref{eqn:higgins-obstacle-cost}  simplifies to $[\nicefrac{r_o}{d_o}(r_{fov}^2 - d_o^2)]^2$ to prevent overflow.  For the remainder of this work, we refer to this method as \emph{Higgins}.

\subsubsection{Andersen et al.~\texorpdfstring{\cite{andersen2019trajectory}}{Andersen}}
In~\cite{andersen2019trajectory}, the authors implement visibility maximization in a study on overtaking maneuvers, where they consider the visibility around an upcoming obstacle. We have implemented the visibility portion of their approach, considering
the closest approaching vehicle only. This contrasts both~\cite{higgins2022model} and the proposed method, which consider all occlusions within range.  We could adjust the visibility cost function (13) to account for multiple occlusions, mainly affecting the scaling parameter $\lambda$; however, we felt that the contrast in a single evaluation would be more interesting. 
 Visibility is maximized by rewarding trajectories that maximize the smallest angle between the AV and the vertices of a bounding box around the closest obstacle $\mathbf{o} \in \mathcal{O}$ along the trajectory.

Let $\phi_o$ be the angle between the critical vertex of the obstacle $\mathbf{o}$ and the AV at $\mathbf{x}_t$.   Then, for each time step, the visibility reward is
\begin{align} \label{eqn:Andersen-obstacle-cost}
    J_{\textsc{vis}}(\mathbf{x}_t) = - \lambda \left( \min_{\mathbf{o} \in \mathcal{O}} \psi_{\mathbf{o},\mathbf{x}_t} \right),
\end{align}
where $\lambda \in \mathbb{R}^+$ is a scaling factor.   We assume that once the AV passes an obstacle, the critical angle has been maximized, and that obstacle can be removed from further consideration.  We refer to this method as \emph{Andersen} in our results.

\subsubsection{Proposed Method}
The \emph{Proposed} method uses a function function $P(\mathbf{x}_t)$ to look up the perspective cost for the cell in $\mathbf{P}_t$ containing $\mathbf{x}_t$.  Using $M$ as a scaling factor, we write the visibility cost function as a reward,
\begin{align}
    J_{\textsc{vis}}(\mathbf{x}_t) = - M P(\mathbf{x}_t).
\end{align}

\subsubsection{No Visibility Cost and Nominal}
As an additional baseline, we include \emph{None}, a method that includes obstacle avoidance only, with no visibility maximization; hence,
\begin{align}
    J_{\textsc{vis}}(\mathbf{x}_t) = 0.
\end{align}
The \emph{Nominal} method implemented MPPI with model constraints only, ignoring obstacles.  This represents the best performance for the given optimization parameters.

\subsection{Weight Tuning}

\begin{figure}
    \centering
    \includegraphics[width=0.5\linewidth]{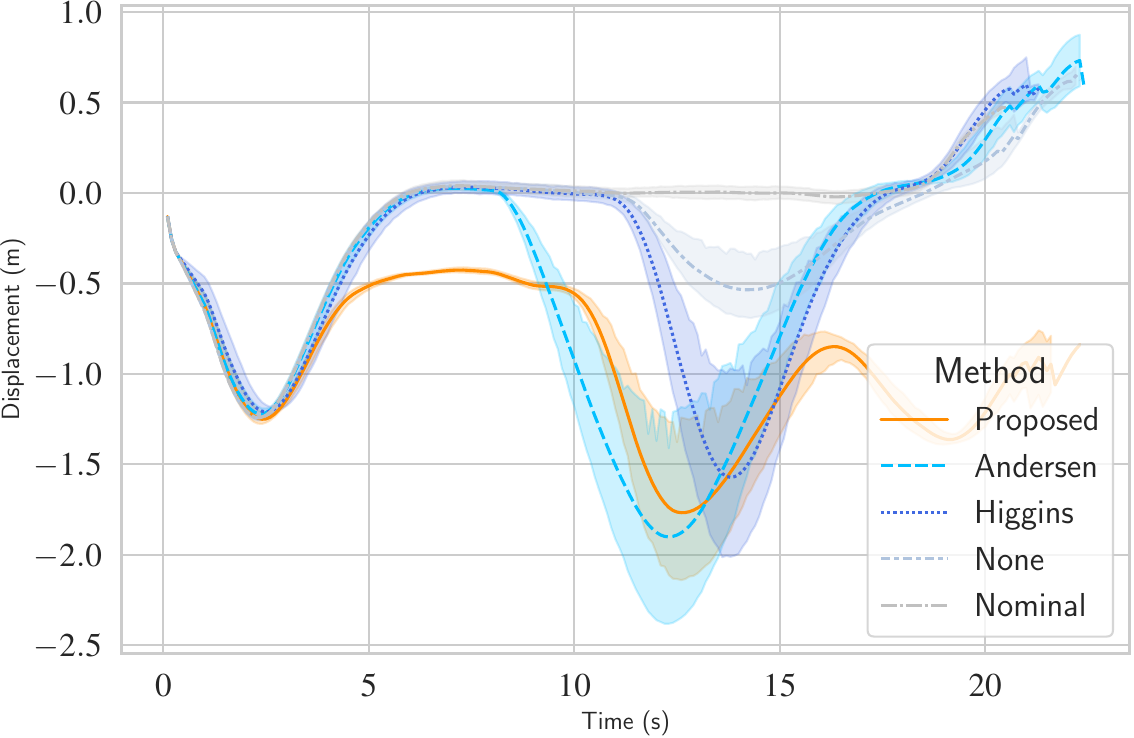}
    \caption{The magnitude of the response to an occlusion is controlled by the weighting parameter $M or \lambda$.  We placed a single vehicle and selected the weight values that resulted in similar responses, creating a basis for equitable comparisons in dense clutter scenarios.}
    \label{fig:single-response}
\end{figure}

To compare methods, we adjusted the weight parameter ($M$ or $\lambda$) to scale the magnitude of the displacement when passing a single car in the \emph{Straight} scenario, targeting a distance of $\sim 2$ m from the nominal path.  We can see from Figure~\ref{fig:single-response} and Table~\ref{tbl:single-obstacle-response} that the methods perform similarly.

%
\begin{table}
\caption{"Mean Displacement and Velocity for a Single Occlusion }
\label{tbl:single-obstacle-response}
\begin{center}
\resizebox{\linewidth}{!}{
\begin{tabular}{@{} l  l  l  c c  c c  c c  @{}}
\toprule
   &  &  & \multicolumn{2}{c}{ Displacement } & \multicolumn{2}{c}{ V } & \multicolumn{2}{c}{ Min Distance }\\
Scenario & Speed & Method & $\mu$ & $\sigma$ & $\mu$ & $\sigma$ & $\mu$ & $\sigma$\\
\midrule
Single Car & 7.5 & Proposed &  -0.9 &  0.49 &   6.7 &  1.16 &   3.5 &  1.87\\
 &  & Andersen &  -0.5 &  0.78 &   6.6 &  1.20 &   3.3 &  1.76\\
 &  & Higgins &  -0.4 &  0.63 &   6.8 &  1.23 &   3.9 &  2.05\\
\bottomrule
\end{tabular}
}
\end{center}
\end{table}
%

\section{Simulation and Discussion}
In this section, we present our experimental results.

\begin{figure*}
    \centering
    \begin{subfigure}{0.24\linewidth}
        \centering
        \includegraphics[width=0.8\linewidth]{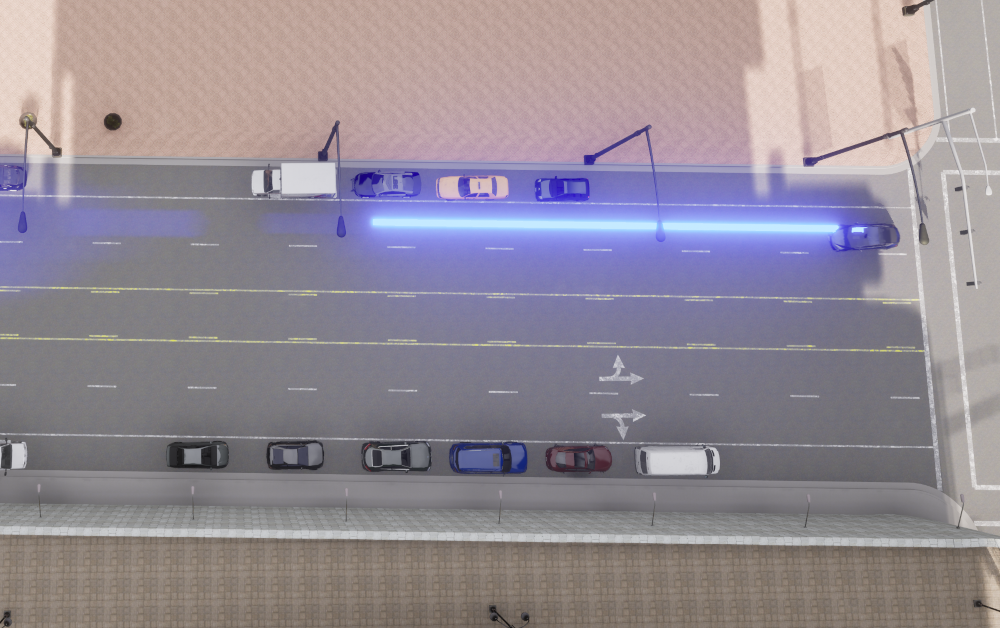}
        \caption{Straight}
        \label{fig:scenario-straight}
    \end{subfigure} 
    \begin{subfigure}{0.24\linewidth}
        \centering
        \includegraphics[width=0.8\linewidth]{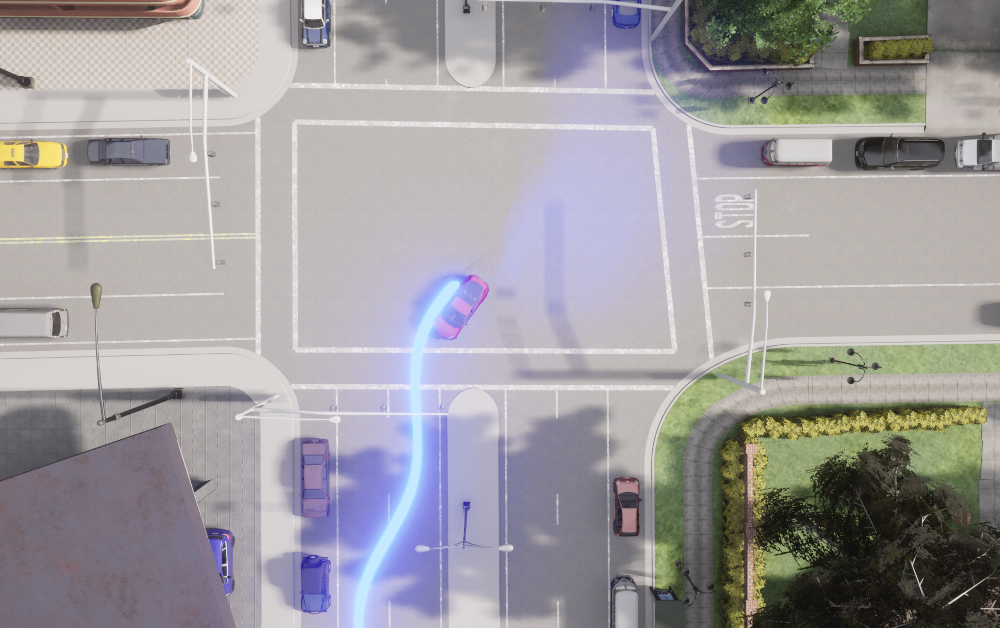}
        \caption{Intersection}
        \label{fig:scenario-intersection}
    \end{subfigure}
    \begin{subfigure}{0.24\linewidth}
        \centering
        \includegraphics[width=0.8\linewidth]{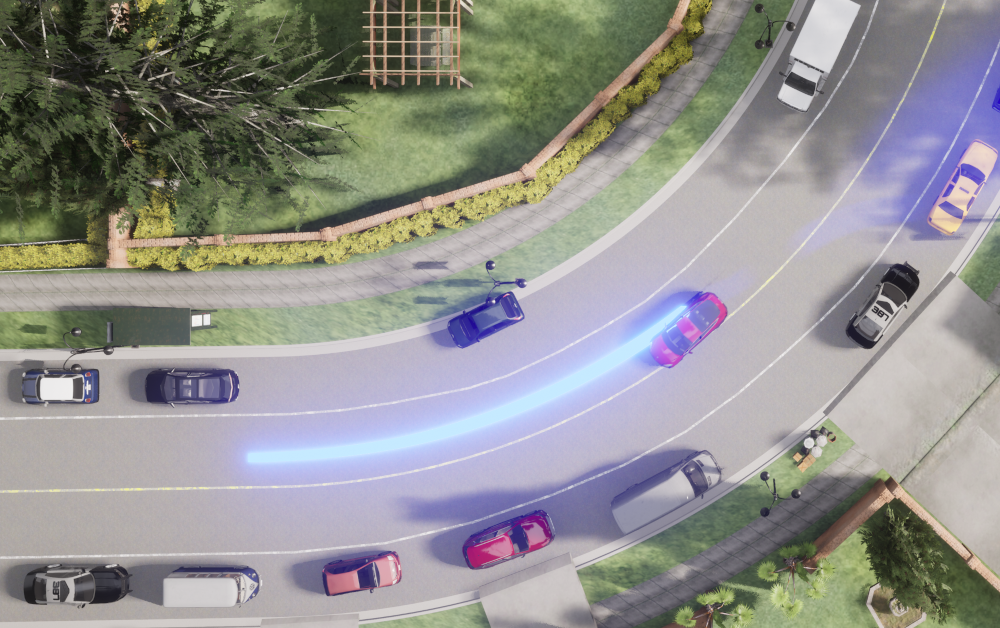}
        \caption{Curve}
        \label{fig:scenario-curve}
    \end{subfigure} 
    \begin{subfigure}{0.24\linewidth}
        \centering
        \includegraphics[width=0.8\linewidth]{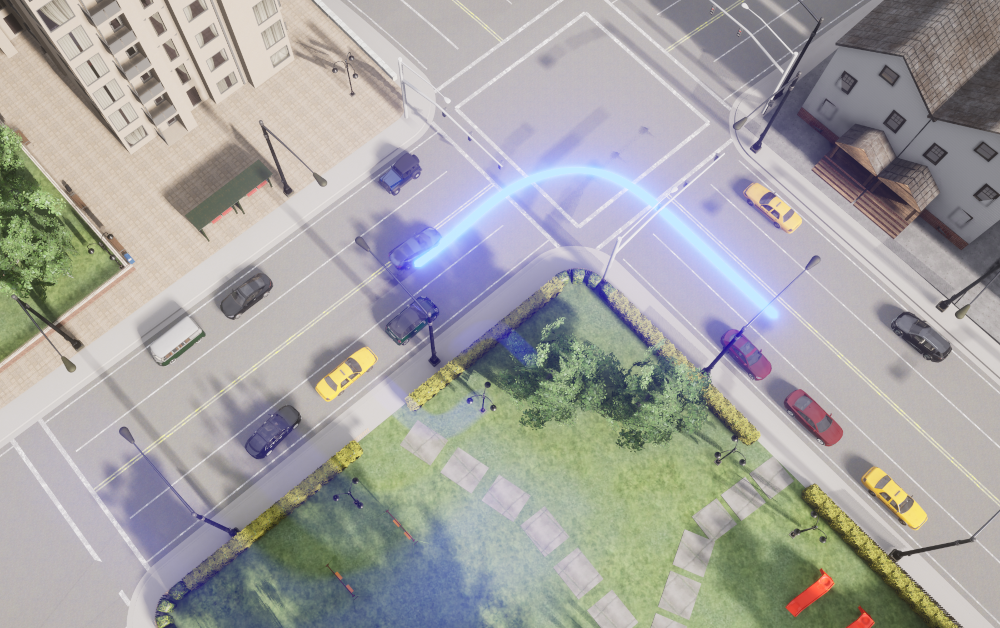}
        \caption{Park}
        \label{fig:scenario-park}
    \end{subfigure} 
    \caption{Experiments are run in one of these four scenarios, each with randomly placed parked cars.  The first two are sparsely cluttered: (\protect\subref{fig:scenario-straight}) a multilane straight section of road, and (\protect\subref{fig:scenario-intersection}) an open intersection. The second two are more densely cluttered: (\protect\subref{fig:scenario-curve})a slow curve, and (\protect\subref{fig:scenario-curve}) a two-lane road outside a park.  A light blue line indicates the nominal path.}
    \label{fig:scenario}
\end{figure*}

We run our experiments in four different scenarios in the CARLA~\cite{dosovitskiy2017carla} open-source simulator environment:  \emph{Straight}, a straight multilane road, \emph{ Intersection}, a wide intersection, \emph{Curve}, a long suburban curve, and \emph{Park}, a two-lane street near a park (see Figure~\ref{fig:scenario}).  
Parked cars are randomly placed along the roads.  The AV follows a trajectory through the scenario at one of three desired speeds, 5.0, 7.5, or 10 m/s.  In our results, we consider the deviation from the nominal trajectory, the actual speed, and the distances from occlusions.  Each experiment is repeated 10 times. 
The MPPI controller samples 10,000 trajectories per time step over a horizon of $25$ steps, with a $0.1$s interval. 
Phantom pedestrians are modeled with $V_{max}^a = 1.9$ m/s, to capture the possibility of a pedestrian running out of an occluded region.  
All simulations are run on a desktop computer with an AMD Ryzen 9 5950X 16-Core Processor and an NVIDIA RTX 3090 GPU.  The underlying cost map is $80$ m x $80$ m at a resolution of $0.4$ m. The mean costmap update required $56$ ms with a standard deviation of $17$ ms. The controller required $2$ ms on average with a $0.5$ ms standard deviation. 

\subsection{Simulation Results}

In our simulations, we measure the AV deviation from the nominal trajectory, the actual velocity, and the minimum distance to obstacles.  The deviation is measured from the desired centerline, with negative values indicating that the car has moved toward the center of the road. Actual velocity measures the AV speed after the actions of the safety controller.  Finally, the minimum distance is the smallest distance to any obstacle over the length of the trajectory. Graphs for experiments at $7.5$ m/s are shown in Figure~\ref{fig:scenario-results} and a summary of all target speeds in Table~\ref{tbl:scenario-data}.

\subsubsection{Earlier Views of Occlusions}

To gain an earlier view of an occlusion, the AV must adjust its trajectory away from the obstacle that causes the occlusion.  As can be seen in Figure~\ref{fig:visibility-demo}, moving away from the obstacle increases the viewing angle and reduces the occluded area.  We see the same behavior emerge when planning with the APCM, as can be seen in Figures~\ref{fig:sparse-displacement-7.5} and~\ref{fig:dense-displacement-7.5}.  A negative deviation indicates that the AV is moving toward the center of the road, away from an approaching obstacle, and towards positions with a higher perspective value.  

\subsubsection{Improved Response to Increased Clutter}

\begin{figure}
    \centering
    \begin{subfigure}{0.4\linewidth}
        \centering
        \includegraphics[width=0.8\linewidth]{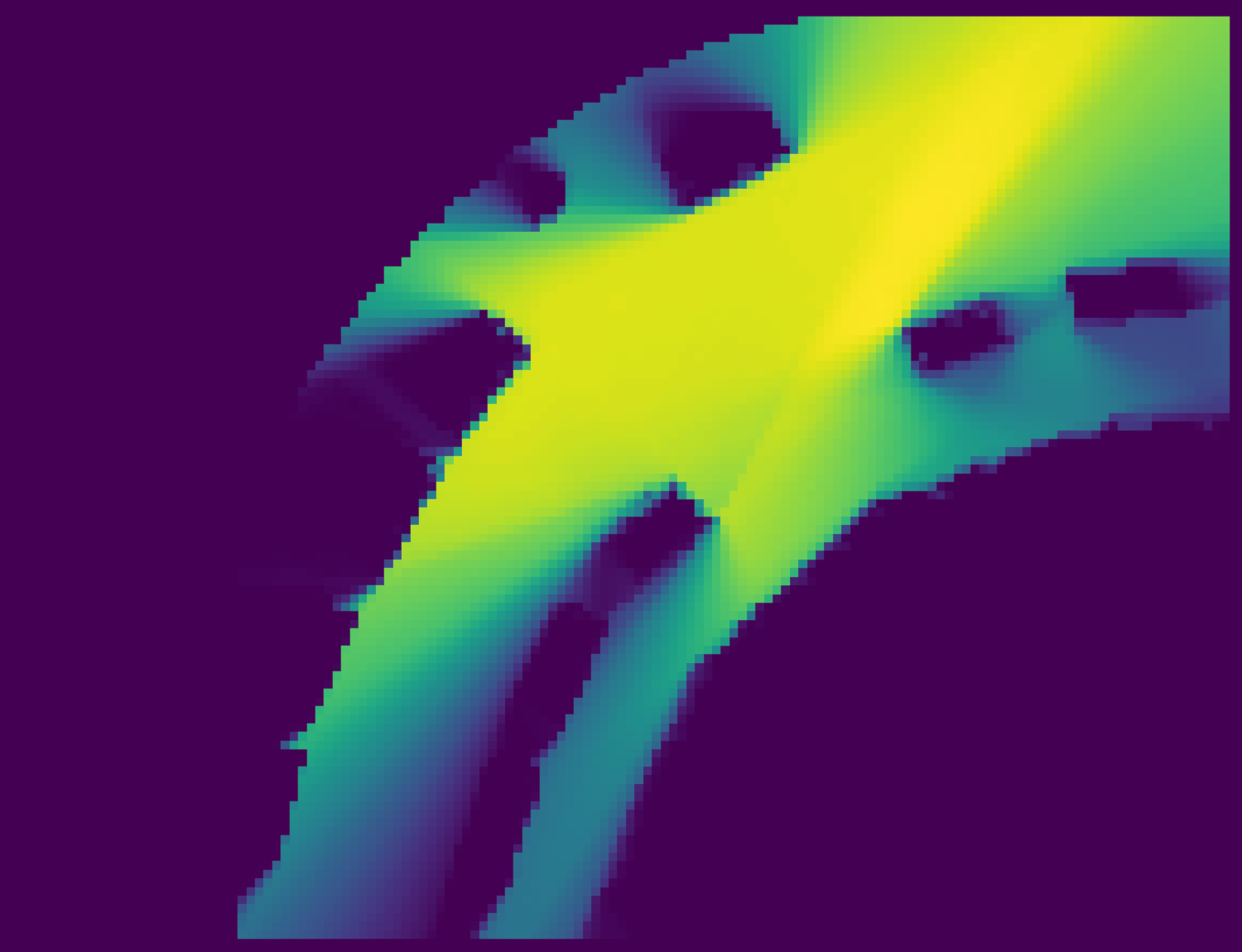}
        \caption{}
        \label{fig:apcm-sample}
    \end{subfigure} 
    \begin{subfigure}{0.4\linewidth}
        \centering
        \includegraphics[width=0.8\linewidth]{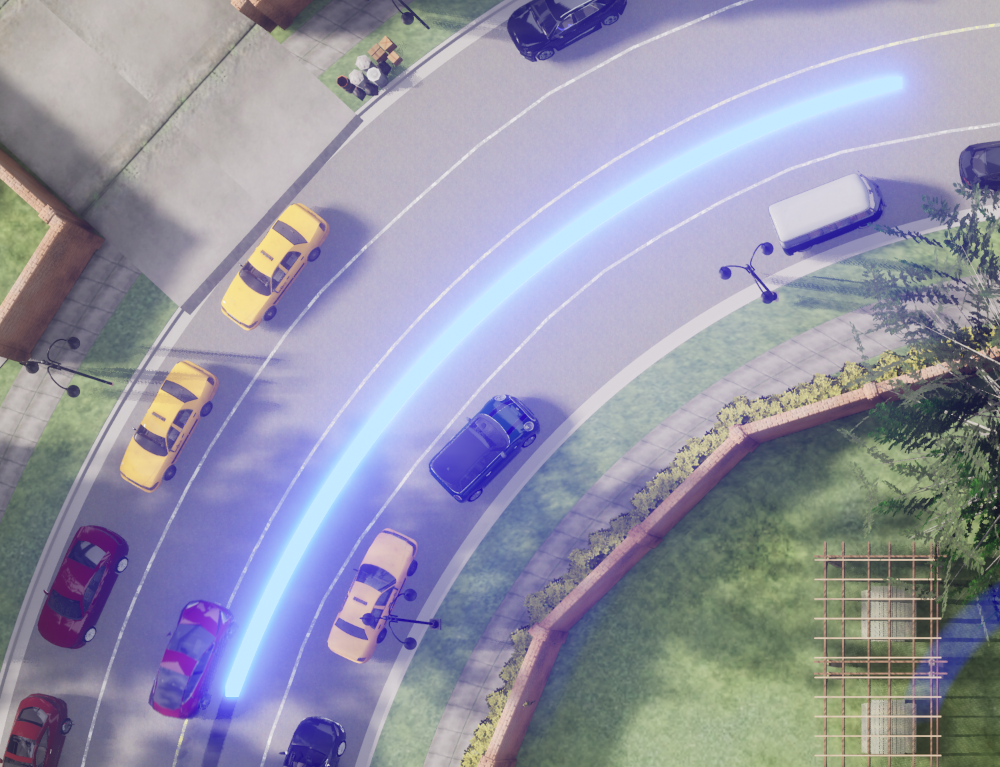}
        \caption{}
        \label{fig:apcm-source}
    \end{subfigure} 
    \caption{An example APCM (\protect\subref{fig:apcm-sample}) drawn from the scenario in (\protect\subref{fig:apcm-source}). The AV is in the bottom left corner.  Brighter cells have a higher probability of viewing $\mathcal{U}_t^r$ and therefore, are expected to result in better perception if visited.}
    \label{fig:false-color-detail}
\end{figure}

%
\begin{table}
\caption{ Mean Displacement and Velocity of the AV for each method. In Sparse clutter scenarios, the methods have similar results, with Higgins having a slight performance edge. 
 In Dense scenarios, the Proposed method maintains a larger minimum distance from obstacles and performs better as a result. }
\label{tbl:scenario-data}
\begin{center}
\resizebox{\linewidth}{!}{
\begin{tabular}{@{} l  l  l  c c  c c  c c  @{}}
\toprule
   &  &  & \multicolumn{2}{c}{ Displacement } & \multicolumn{2}{c}{ V } & \multicolumn{2}{c}{ Min Distance }\\
Clutter & Speed & Method & $\mu$ & $\sigma$ & $\mu$ & $\sigma$ & $\mu$ & $\sigma$\\
\midrule
Sparse & 5.0 & Proposed &  \textbf{-0.8} &  \textbf{0.85} &   \textbf{4.6} &  \textbf{0.81} &   4.5 &  0.38\\
 &  & Andersen &  -0.7 &  1.22 &   4.6 &  0.80 &   4.3 &  0.52\\
 &  & Higgins &  \textbf{-0.8} &  \textbf{1.27} &   4.5 &  0.81 &   \textbf{4.8} &  \textbf{0.46}\\
 \cmidrule{2-9}
 & 7.5 & Proposed &  \textbf{-0.9} &  \textbf{0.86} &   6.8 &  1.06 &   4.3 &  0.38\\
 &  & Andersen &  -0.7 &  1.16 &   6.7 &  1.10 &   4.2 &  0.46\\
 &  & Higgins &  -0.8 &  1.18 &   \textbf{6.9} &  \textbf{1.01} &   \textbf{4.8} &  \textbf{0.48}\\
 \cmidrule{2-9}
 & 10.0 & Proposed &  \textbf{-0.8} &  \textbf{0.87} &   7.4 &  1.50 &   4.1 &  0.57\\
 &  & Andersen &  -0.7 &  1.19 &   7.4 &  1.61 &   4.0 &  0.60\\
 &  & Higgins &  \textbf{-0.8} &  \textbf{1.19} &   \textbf{7.6} & \textbf{ 1.52} &   \textbf{4.6} &  \textbf{0.65}\\
\midrule
Dense & 5.0 & Proposed &  \textbf{-0.8} &  \textbf{0.61} &   \textbf{4.6} &  \textbf{0.83} &   \textbf{4.8} &  \textbf{0.23}\\
 &  & Andersen &  -0.3 &  0.89 &   4.5 &  0.84 &   3.5 &  0.42\\
 &  & Higgins &  -0.5 &  0.70 &   4.5 &  0.87 &   4.3 &  0.29\\
 \cmidrule{2-9}
 & 7.5 & Proposed &  \textbf{-1.0} &  \textbf{0.82} &   \textbf{6.6} &  \textbf{1.11} &   \textbf{4.5} &  \textbf{0.20}\\
 &  & Andersen &  -0.3 &  0.78 &   6.1 &  1.47 &   3.4 &  0.46\\
 &  & Higgins &  -0.4 &  0.62 &   6.3 &  1.35 &   4.3 &  0.26\\
 \cmidrule{2-9}
 & 10.0 & Proposed &  \textbf{-0.9} &  \textbf{0.78} &   \textbf{7.1} &  \textbf{1.51} &   \textbf{4.3} &  \textbf{0.18}\\
 &  & Andersen &  -0.1 &  0.85 &   6.3 &  2.08 &   3.2 &  0.40\\
 &  & Higgins &  -0.3 &  0.65 &   6.5 &  1.99 &   4.0 &  0.28\\
\bottomrule
\end{tabular}
}
\end{center}
\end{table}
%

Our simulation results demonstrate an improved response as clutter in the scenario increases.  We define clutter in a scenario by the number of occlusions and their proximity as the AV follows a trajectory.  We used the mean of the minimum distances between the AV and each obstacle as an approximate measure of clutter.  Scenarios \emph{Straight} and \emph{Intersection} have sparse clutter, with a mean minimum distance to the AV trajectory of 12 m and a standard deviation of 6 m.  The \emph{Curve} and \emph{Park} scenarios have dense clutter with a mean minimum distance of 6 m and a standard deviation of $0.3$ m.

\begin{figure*}
    \centering
    \begin{subfigure}{0.24\linewidth}
        \includegraphics[width=0.8\linewidth]{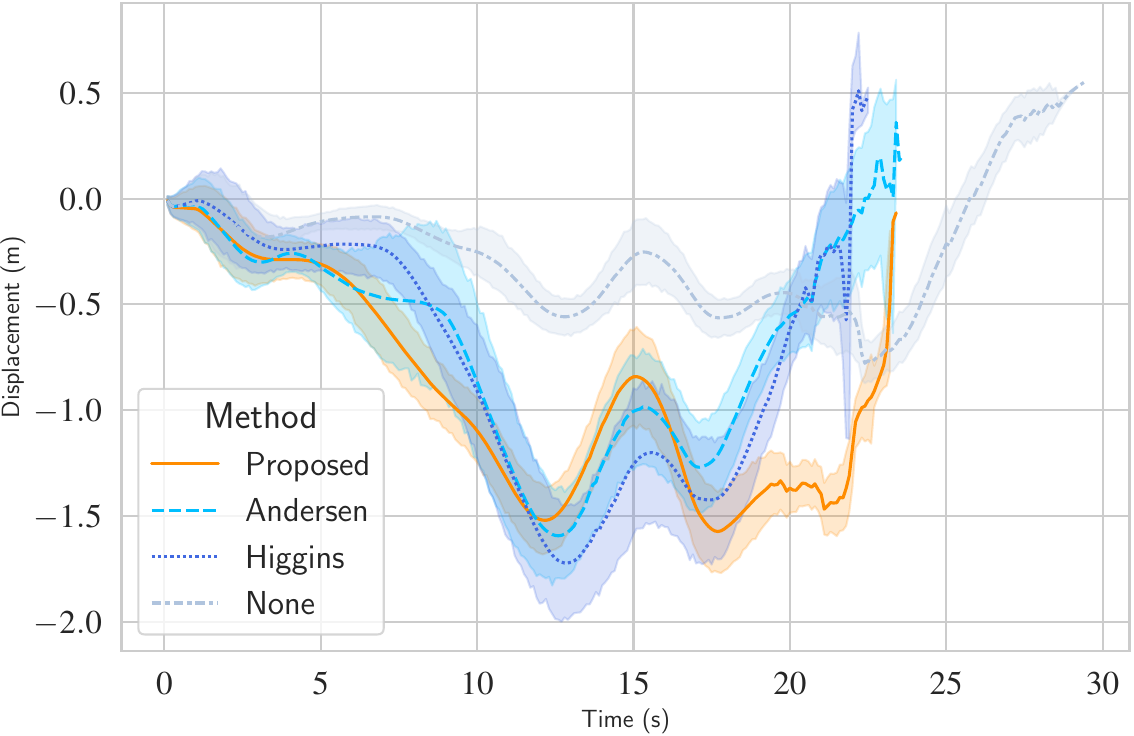}
        \caption{Sparse (7.5 m/s)}
        \label{fig:sparse-displacement-7.5}
    \end{subfigure} 
    \begin{subfigure}{0.24\linewidth}
        \includegraphics[width=0.8\linewidth]{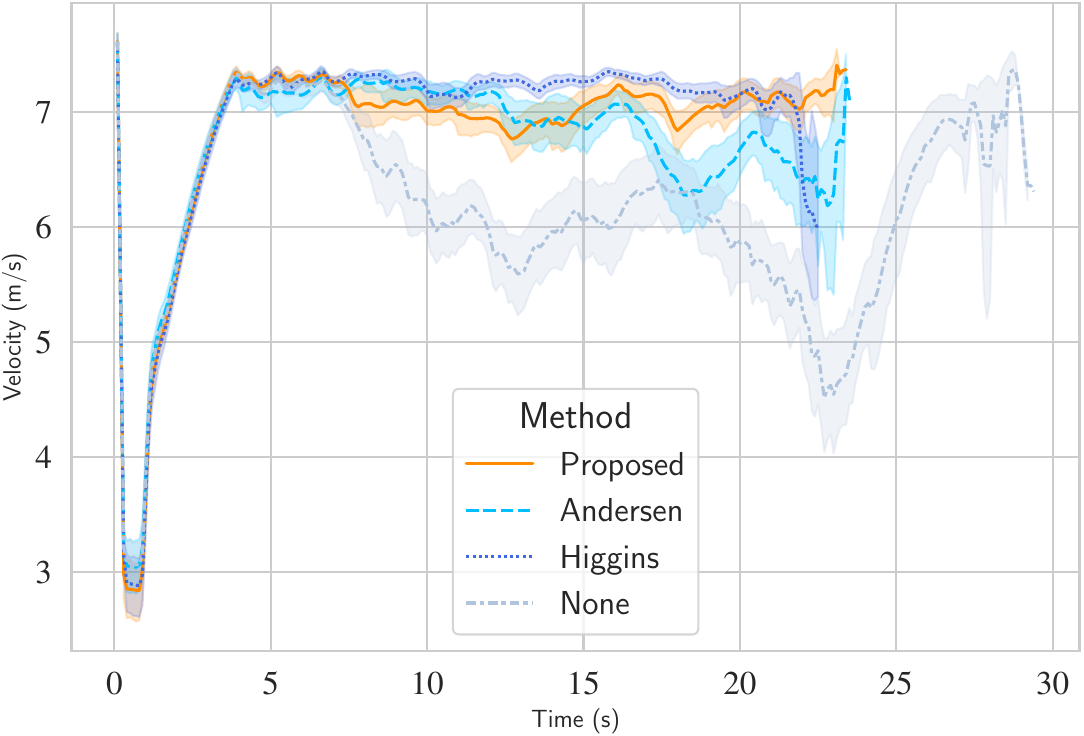}
        \caption{Sparse (7.5 m/s)}
        \label{fig:sparse-speed-7.5}
    \end{subfigure} 
    \begin{subfigure}{0.24\linewidth}
        \includegraphics[width=0.8\linewidth]{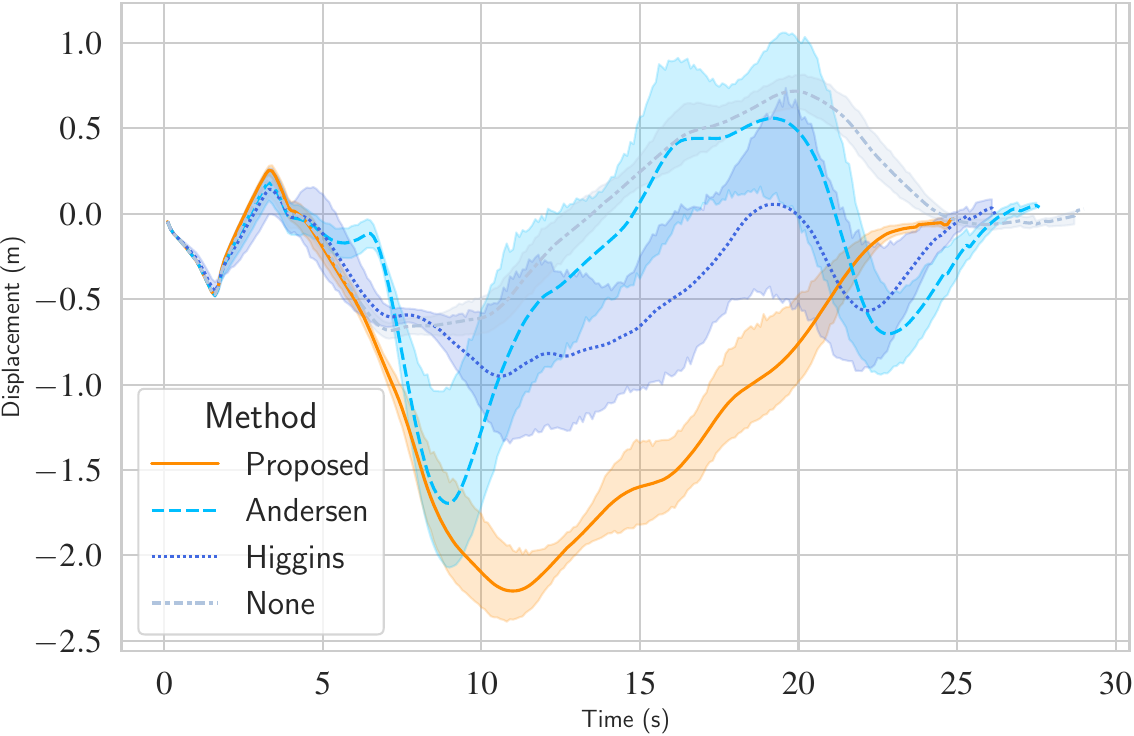}
        \caption{Dense (7.5 m/s)}
        \label{fig:dense-displacement-7.5}
    \end{subfigure} 
    \begin{subfigure}{0.24\linewidth}
        \includegraphics[width=0.8\linewidth]{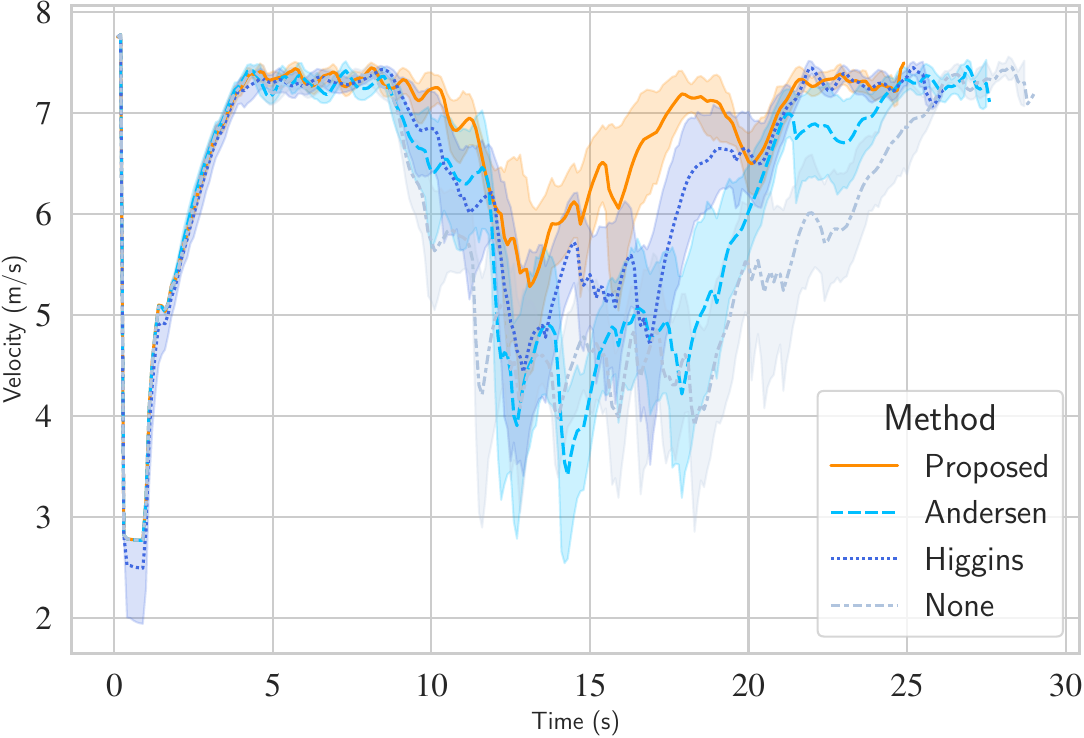}
        \caption{Dense (7.5 m/s)}
        \label{fig:dense-speed-7.5}
    \end{subfigure}
    \caption{Comparing trajectories in scenarios with sparse \protect\subref{fig:sparse-displacement-7.5},\protect\subref{fig:sparse-speed-7.5}) and dense clutter (\protect\subref{fig:dense-displacement-7.5}-\protect\subref{fig:dense-speed-7.5}).  In sparse scenarios, all methods are similar, with Higgins having a slight edge. In dense scenarios, the Proposed method has a higher average velocity due to its ability to maintain a larger displacement from nominal.}
    \label{fig:scenario-results}
\end{figure*}

In sparse clutter scenarios, all visibility planners shift the AV trajectory toward the center of the road as expected, resulting in more consistent speeds, as can be seen in Figures~\ref{fig:sparse-speed-7.5}, and~\ref{fig:dense-speed-7.5}.  All of the planners, with the exception of None, perform similarly, showing only minor differences.   Predictably, the obstacle-only baseline, None, is much slower, as it does not deviate enough to maintain its speed safely.

In dense clutter scenarios, the Proposed method is able to successfully combine the conflicting cost, as shown in Figure~\ref{fig:dense-displacement-7.5} where the Proposed method maintains a higher deviation.  This can also be seen in Table~\ref{tbl:scenario-data} where the Proposed method maintains a larger minimum distance to occlusions along the trajectory in dense scenarios.  

We intuit this is the result of how the optimizing cost function is used. Andersen and Higgins use the optimizer cost function to directly sum the demands caused by obstacles.
When the clutter in the scenario is sparse, the resulting visibility costs tend to push the AV away from the occlusions.  This reduces the area of occlusions that affect the AV, reducing the possibility of a collision with a hidden pedestrian. 
When the scenario has dense clutter, multiple occlusions create conflicting costs, which when summed in the cost function, reduce or even eliminate the desired response.
The result can be seen in Figure~\ref{fig:dense-displacement-7.5}.  Both Higgins and Andersen have a reduced deviation response, similar to None.   
However, the Proposed method performs the perception summation in the cost map. As a result, costs are not conflicting as in baselines, but are additive, resulting in better performance in dense clutter.   This effect is illustrated in Figure~\ref{fig:false-color-detail}, where cells near the center of the road have a higher value because they have a higher probability of seeing more of $\mathcal{U}_t^r$.  By resolving competing demands on the APCM, we achieve a better result than relying solely on the OCP cost function as in the existing methods.

\begin{remark} [Deviation of None Method]
    Note that the None method also deviates slightly despite not having visibility maximization.  This deviation is a side effect of the sampling method employed by MPPI, where the sampled trajectories that intersect an obstacle are heavily penalized.  This results in a slight bias that causes a deviation from the nominal, although not enough to avoid slowing.  
\end{remark}

\section{Conclusion}

In this work, we address the problem of evaluating future visibility as a function of an alternate perspective cost map.  We showed that by constructing a cost map based on alternate perspectives, we enable visibility-based planning environment with dense clutter.  We demonstrated our solution using an MPC-based planner in simulation, where we showed how the cost map approach successfully resolved conflicting visibility demands.   Finally, we released a GPU-based reference implementation to illustrate our claims.

Although this paper has focused on vehicle path planning, APCM could extend to other applications where robots need to consider occlusions when planning.  For example, small delivery robots must work in pedestrian-cluttered environments or in robot trackers that must maintain line-of-sight contact with a target. In all of these applications, we believe that a cost map would enable a planner to anticipate more information-rich trajectories.

One of the limitations of this work is that as the resolution of the cost map decreases, the computation time can increase dramatically.  By replacing the dense observation grid with a sparse representation, we expect to be able to support much finer cost maps.  Furthermore, $\mathcal{U}_t^r$ is calculated only for the initial time and assumes a static state limiting the accuracy of some perception predictions.  For future work, we plan to incorporate a dynamic occupancy grid that allows the generation of a more accurate APCM when there are moving actors in the scene.  Finally, we plan to extend our methods to recorded data sets to better explore real-world performance.

\bibliographystyle{IEEEtran}
\bibliography{references}

\end{document}